\documentclass[a4paper,twoside]{article}

\usepackage{epsfig}
\usepackage{subfigure}
\usepackage{calc}
\usepackage{amssymb}
\usepackage{amstext}
\usepackage{amsmath}
\usepackage{amsthm}
\usepackage{multicol}
\usepackage{pslatex}
\usepackage{apalike}
\usepackage{SCITEPRESS}     

\usepackage{times}
\usepackage{latexsym}
\usepackage{url}
\usepackage{graphicx}
\usepackage{amsmath}
\usepackage{subfig}

\begin{document}

\title{Towards Controlled Transformation of Sentiment in Sentences}

\author{\authorname{Wouter Leeftink and Gerasimos Spanakis}
\affiliation{Department of Data Science and Knowlednge Engineering, Maastricht University\\Maastricht, 6200MD, Netherlands}
\email{w.leeftink@student.maastrichtuniversity.nl, jerry.spanakis@maastrichtuniversity.nl}
}


\abstract{
An obstacle to the development of many natural language processing products is the vast amount of training examples necessary to get satisfactory results. The generation of these examples is often a tedious and time-consuming task. This paper this paper proposes a method to transform the sentiment of sentences in order to limit the work necessary to generate more training data. This means that one sentence can be transformed to an opposite sentiment sentence and should reduce by half the work required in the generation of text. The proposed pipeline consists of a sentiment classifier with an attention mechanism to highlight the short phrases that determine the sentiment of a sentence. Then, these phrases are changed to phrases of the opposite sentiment using a baseline model and an autoencoder approach. Experiments are run on both the separate parts of the pipeline as well as on the end-to-end model. The sentiment classifier is tested on its accuracy and is found to perform adequately. The autoencoder is tested on how well it is able to change the sentiment of an encoded phrase and it was found that such a task is possible. We use human evaluation to judge the performance of the full (end-to-end) pipeline and that reveals that a model using word vectors outperforms the encoder model. Numerical evaluation shows that a success rate of 54.7\% is achieved on the sentiment change.
}

\onecolumn \maketitle \normalsize \vfill

\section{\uppercase{Introduction}}
In its current state text generation is not able to capture the complexities of human language, making the generated text often of poor quality. \cite{hu2017toward} suggested a method to control the generation of text combining variational autoencoders and holistic attribute discriminators. Although the sentiment generated by their method was quite accurate, the generated sentences were still far from perfect. The short sentences generated by their model seem adequate, but the longer the sentence, the more the quality drops.

Most research tries to generate sentences completely from scratch and while this is one way to generate text, it might also be a possibility to only change parts of a sentence to transform the sentiment. In longer sentences, not every word is important for determining the sentiment of the sentence, so most words can be left unchanged while trying to transform the sentiment.

The model proposed in this work tries to determine the critical part of a sentence and transforms only this to a different sentiment. This method should change the sentiment of the sentence while keeping the grammatical structure and semantic meaning of the sentence intact. To find the critical part of a sentence the model uses an attention mechanism on a sentiment classifier. The phrases that are deemed important by the sentiment classifier are then encoded in an encoder-decoder network and transformed to a new phrase. This phrase is then inserted in the original sentence to create the new sentence with the opposite sentiment. 

\section{\uppercase{Related work}}
Word embeddings \cite{word2vec,pennington2014glove} are often used in Natural language processing tasks as they capture the semantic information of words. Both the glove and word2vec algorithm for word embeddings are based on the old idea of distribution hypothesis \cite{firth1957synopsis}, \cite{harris2016words} which states that words that occur on the same place in a sentence are likely to have a similar meaning and has been re-discovered recently \cite{bengio2009learning}. Word2vec can be trained using the skip-gram or the continuous bag of words (CBOW) approach. The CBOW approach aims to maximize the probability that a word occurs given its context words, whereas the skip-gram approach tries to predict surrounding words given a single word. Glove builds a co-occurence matrix of words which is then used to find which words are similar to eachother.

Sentiment analysis is a task in NLP that aims to predict the sentiment of a sentence \cite{liu2012sentiment}. The task can range from a binary classification task where the aim is to predict whether a document is positive or negative to a fine-grained task with multiple classes. In sentiment analysis, state-of-the-art results have been achieved using neural network architectures such as convolutional neural networks \cite{kim2014convolutional} and recurrent neural networks \cite{tang2015document}. Variants of RNNs; LSTMs and GRUs, have also been used to great success\cite{cho2014learning}.  

The attention mechanism was first proposed for the task of machine translation \cite{bahdanau2014neural}. Attention allows a network to 'focus' on one part of the sentence at a time. This is done through keeping another vector which contains information on the impact of individual words. Attention has also been used in other tasks within NLP area such as document classification \cite{yang2016hierarchical}, sentiment analysis \cite{wang2016attention} and teaching machines to read \cite{machinereading}

Encoder-decoder networks \cite{sutskever2014sequence,cho2014learning} are often used in neural machine translation to translate a sequence from one language to another. These networks use RNNs or other types of neural networks to encode the information in the sentence and the another network to decode this sequence to the target language. Since RNNs do not perform well on longer sequences, the LSTM \cite{sutskever2014sequence} unit is often used for their memory component. Gated Recurrent Units \cite{cho2014learning} are simpler variants on the LSTM, as they do not have an output gate.

Transforming the sentiment of sentences has not been systematically attempted, however there are some previous pieces of research into this particular topic. \cite{DBLP:journals/corr/abs-1804-06437} propose a method where a sentence or phrase with the target attribute, in this case sentiment, is extracted and either inserted in the new sentence or completely replacing the previous sentence. Their approach finds phrases based on how often they appear in text with a certain attribute and not in text with the other attribute. However, this approach can not take phrases into account that by themselves are not necessarily strongly leaning towards one sentiment, but still essential to the sentiment of the sentence. 

\cite {larssonmanifold} propose a method that uses manifold traversal to move the sentiment of a sentence from one sentiment to another. Their approach does not keep any of the initial sequence intact and often produces output that is of low quality. The model uses a encoder-decoder architecture with a CNN as encoder and RNN as decoder, optimizing towards the sentiment of the text.

\section{\uppercase{The model}}
\begin{figure*}[ht]
\includegraphics[scale=0.49]{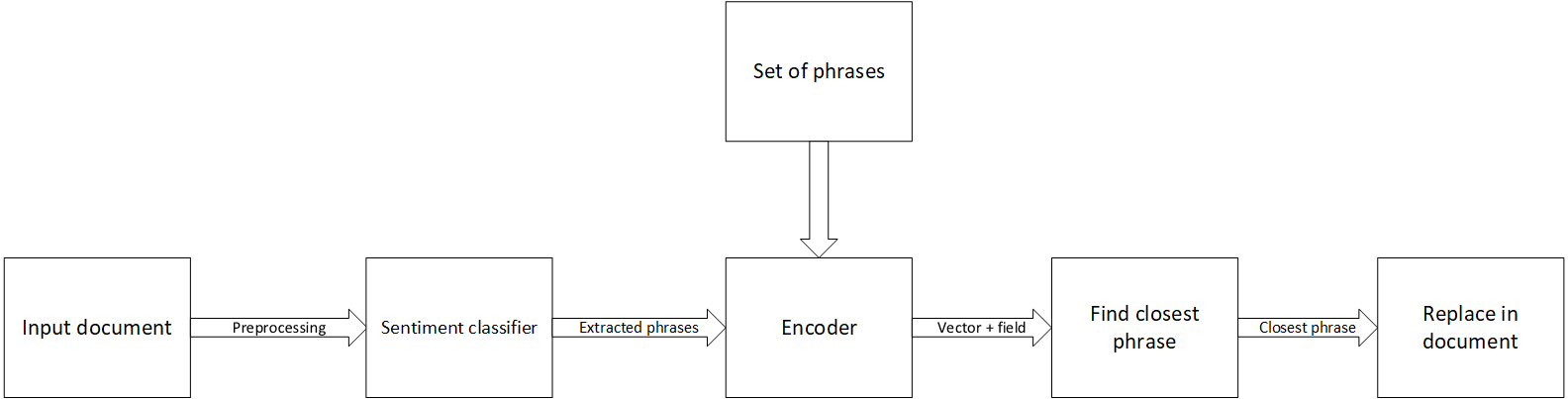} 
\includegraphics[scale=0.58]{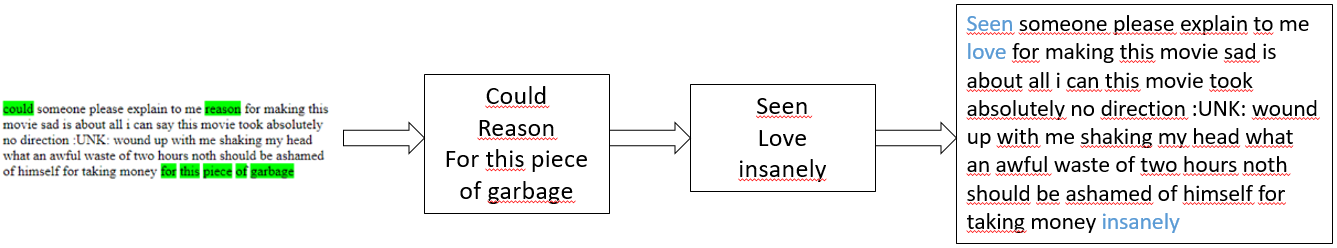}
\caption{The model using the closest phrase approach including an example}
\label{fig:closephrase}
\end{figure*}

In this paper two different pipelines are considered. Both pipelines contain a sentiment classifier with an attention mechanism to extract phrases from the input documents. The difference is that pipeline 1 (which can be seen in Figure \ref{fig:closephrase}) uses an encoder to encode the extracted phrases and find the closest phrase with the opposite sentiment in the vector space. This phrase is then either inserted into the sentence or the vector representation of this phrase is decoded and the resulting phrase is inserted into the sentence. Pipeline 2 (as seen in Figure \ref{fig:wordvec}) finds the words in the extracted phrases that are most likely to determine the sentiment and replaces these words with similar words of the opposite sentiment using word vectors. In the next sections all individual parts of the pipeline will be explained. 
\subsection{Sentiment classification with attention}
To find the phrases that determine the sentiment of a sentence a sentiment classification model with attention is used. The network used is the network defined by \cite{yang2016hierarchical}. This model is chosen because in sequence modeling recurrent neural networks have shown to give better classification results than other models, such as convolutional neural networks \cite{yin2017comparative}. Recurrent neural networks have the added benefit of easily allowing for implementation of attention mechanisms, which are able to focus on small parts of a sequence at a time. The attention mechanism is used to extract the sequences that determine the sentiment.

This classifier consists of a word- and sentence encoder and both a word and sentence level attention layer.  The word encoder is a bidirectional GRU that encodes information about the whole sentence centered around word $w_{it}$ with $t \in [1,T]$. The sentence encoder does the same thing, but for a sentence $s_i$ which is constructed by taking an aggregate of the word vectors and the attention values composing the sentence. The sentence is then encoded by another bidirectional GRU and another attention layer to compose a document vector. This document vector can then be used to determine the sentiment of the document through the fully-connected layer. 

\subsubsection{Attention for phrase extraction}
The attention proposed by \cite{yang2016hierarchical} is used to find the contribution to the sentiment that each individual word brings. The attention they propose feeds the word representation obtained from the word-level encoder through a one-layer MLP to get a hidden representation of the word. This hidden representation is then, together with a word context vector, fed to a softmax function to get the normalized attention weight. After the attention weights have been computed, words with a sufficiently high attention weight are extracted and passed to the encoder-decoder model.

\begin{figure*}[t]
\includegraphics[scale=0.49]{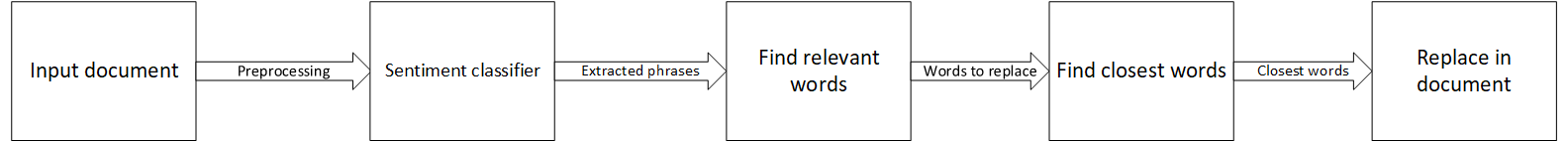}
\includegraphics[scale=0.58]{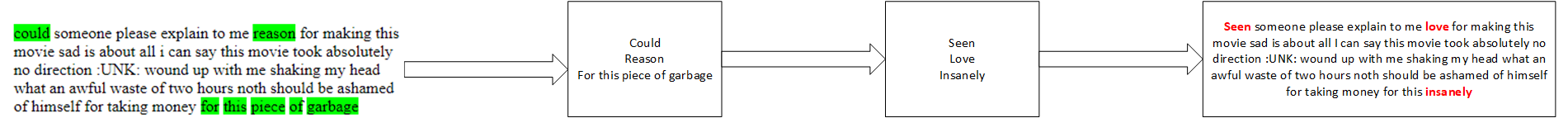}
\caption{The model using the word vector approach including an example}
\label{fig:wordvec}
\end{figure*}

In the first step of the examples in Figure \ref{fig:closephrase} and \ref{fig:wordvec} the attention mechanism is highlighting a few phrases in a movie review. These are the phrases that are then later passed on to either the encoder or to the word changing part of the pipeline.

\subsection{Sentiment transformation}
For the sentiment transformation two approaches are proposed. The first approach is based on an encoder which encodes the extracted phrases into fixed-length vectors. The second approach transforms words from the extracted phrases using word embeddings and an emotion lexicon.

\subsubsection{Encoder-Decoder approach}
The encoder is the first technique used to transform a sentence to one with a different sentiment. This variant uses an encoder model and a transformation based on the distance between two phrases in the latent space. The first part of this model is to encode the phrases extracted by the attention in the latent space. The encoder used is similar to that proposed by \cite{cho2014learning}. 

The difference is that this model is not trained on two separate datasets, but on one set of phrases both as input and output, where the goal is that the model can echo the sequence. Both the encoder and the decoder are one-directional GRUs that are trained together to echo the sequence. First, the encoder encodes a sequence to a fixed-length vector representation. The decoder should then reconstruct the original sequence from this fixed-length vector representation. This is trained through maximizing the log-likelihood \[\max_{\theta} \frac{1}{N}\sum_{n=1}^{N}log p_\theta(y_n \mid x_n)\] where $\theta$ is the set of model parameters and $(x_n, y_n)$ is a pair of input and output sequences from the training set. In our case $x_n$ and $y_n$ are the same as we want the encoder-decoder to echo the initial sequence. On top of this we also store a sentiment label with the encoded sequences to use them in the next step of the model.

Afterwards these phrases are encoded into a fixed-length vector. The model then selects the vector closest to the current latent, fixed-length representation (but taking the one with a different sentiment label) using the cosine distance:
\[\min_\mathbf{y} \frac{\mathbf{x} \cdot \mathbf{y}}{||\mathbf{x}|| \cdot ||\mathbf{y}||}, \mathbf{y} \in Y\]

where $\mathbf{x}$ is the encoded input phrase and $Y$ is the set of all encoded vectors in the latent space with the opposite label from $\mathbf{x}$. 
The closest vector is then decoded into a new phrase, which is inserted into the sentence to replace the old phrase. Obviously, the decoder model used here is the same model which is trained to encode the selected phrases. 
 


\subsubsection{Word vector approach}
The word vector approach also starts by extracting the relevant phrases from the document using the attention mechanism explained in the attention section. However, while the encoder approach uses an encoder to encode the phrases into the latent space, this approach is based on word vectors \cite{word2vec}. First off, the words that are important to the sentiment are selected using the following formulas: \[  \forall x \in X: p(neg | x) > 0.65 \vee p(pos | x) > 0.65\] where $neg$ means the sentiment of the sentence is negative, $pos$ means the sentiment of the sentence is positive, $x$ is the current word and $X$ is the set of all the words selected to be replaced. Threshold $0.65$ was chosen empirically by inspecting different values. 

The replacement word is selected using the closest word in the latent space using the cosine distance. The candidates to replace the word are found using the EmoLex emotion lexicon \cite{Mohammad13}. A negative and a positive word list are created based on this lexicon using the annotations. The negative list contains all words marked as negative and the positive list contains all words marked as positive. When a phrase is positive the closest word in the negative list is chosen and vice-versa when the phrase is negative. The chosen words are then replaced and the new phrase is inserted into the original sentence. Combining both the attention mechanism and the word embeddings was done because it was much faster than going through the whole sequence and replacing words according to the same formula. 

\section{\uppercase{Data}}
The data used come from the large movie review dataset \cite{maas-EtAl:2011:ACL-HLT2011}. This dataset consists of a training set containing 50000 unlabeled, 12500 positive and 125000 negative reviews and a test set containing 12500 positive and 12500 negative reviews. The experiments in this paper were performed only using the positive and negative reviews, which meant the training set contained 25000 reviews and the test set also contained 25000 reviews.

In terms of preprocessing the text was converted to lower case and any punctuation was removed. Lowercasing was done to avoid the same words being treated differently because they were at the beginning of the sentence and punctuation was removed so that the punctuation would not be included in tokens or treated as its own token.

To see if the model would transfer well to another dataset the experiments were repeated on the Rotten tomato review dataset \cite{pang2005seeing}. This dataset was limited to only full sentences and the labels were changed to binary classification labels. Only the instances that were negative and positive were included and the instances that were somewhat negative or somewhat positive (labels 1, 2 and 3) were ignored.

\section{\uppercase{Experiments}}
In order to properly test the proposed method, experiments were ran on both the individual parts of the approach and on the whole (end-to-end) pipeline. Evaluating the full pipeline was difficult as different existing metrics seemed insufficient because of the nature of the project. For example the BLEU-score would always be high since most of the original sequence is left intact and it has been criticized in the past \cite{novikova2017we}. The percentage of sentences that changed sentiment according to the sentiment classifier was used as a metric, but as sentiment classifiers do not have an accuracy of 100\%, this number is a rough estimate. Lastly, a random subset of $15$ sentences was given to a test group of $4$ people and asked whether they deemed the sentences correct and considered the sentiment changed. 

\subsection{Sentiment classifier}
To test the performance of the sentiment classifier individually, the proposed attention RNN model was trained on the 25000 training reviews of the imdb dataset. The sentiment classifier was then used to predict the sentiment on 2000 test reviews of the same dataset. These 2000 reviews were randomly selected. The accuracy of the sentiment classifier was tested because the classifier will later be used in testing the full model and the performance of the encoder-decoder, which makes the performance of the sentiment classifier important to report.

The attention component, for which this sentiment classification model was chosen is more difficult to test. The performance in the attention will be tested by the experiments with the full model. The higher the score for sentiment change is, the better the attention mechanism will have functioned as for a perfect score the attention mechanism will need to have picked out all phrases that contribute towards the sentiment.

The parameters we use consist of an embedding dimension of $300$, a size of the hidden layer of $150$, an attention vector of size $50$ and a batch size of $256$. The network makes use of randomly initialized word vectors that are updated during training. The network is trained on the positive and negative training reviews of the imdb dataset and the accuracy is measured using the test reviews. Loss is determined using cross entropy and the optimizer is the Adam opimizer \cite{kingma2014adam}.  

The proposed sentiment classifier was tested in terms of accuracy on the imdb dataset and in comparison to state of the art models. The numbers used to compare the results are reported by \cite{DBLP:journals/corr/abs-1708-00107}.

\begin{table}
\begin{center}
\begin{tabular}{l|c}
Model & Accuracy \\
\hline
This Model & 89.6\\
SA-LSTM \cite{dai2015semi} & 92.8\\
bmLSTM \cite{radford2017learning} & 92.9\\
TRNN \cite{DBLP:journals/corr/DiengWGP16} & 93.8\\
oh-LSTM \cite{johnson2016supervised} & 94.1\\
Virtual \cite{miyato2016adversarial} & 94.1
\end{tabular}
\end{center}
\caption{Accuracy of the sentiment classifier compared to the state of the art}
\label{tab:sentacc}
\end{table}

Table \ref{tab:sentacc} shows that the result of the sentiment classifier used in this paper is slightly below the state of the art. On the imdb dataset we achieve an accuracy (on a binary sentiment classification task) of 89.6 percent, a bit lower than the state of the art on the same dataset. However, the reason this algorithm is used is its ability to highlight the parts of the sentence that contribute most towards the sentiment, which only the bmLSTM algorithm does. Still, the performance of the classifier is satisfactory enough for being deployed into the full model.

\subsection{Autoencoder}
The autoencoder's purpose is to encode short phrases in the latent space so that the closest phrase of the opposite class (sentiment) can be found. To test the performance of the autoencoder for the task presented in this paper, phrases were extracted from the test reviews of the imdb dataset and were then encoded using the autoencoder. The closest vector was then decoded and the sentiment of the resulting sequence was determined using the sentiment classifier described in this paper. In the results section the percentage of phrases that changed sentiment is reported. This experiment which assesses the the performance of the autoencoder is conducted to better interpret the results of the full model. 

For training, an embedding dimension of $100$ and a size of the hidden layer of $250$ are used. The word vectors used are pretrained GloVe embedding vectors from the GloVe vector set. The network is trained on the training set of phrases acquired by the attention network, using a negative log likelihood loss function and as an optimizer a stochastic gradient optimizer with a learning rate of 0.01 is used. The training objective is to echo the phrases in the training set. After encoding the sentiment label of the phrase is saved together with the fixed-length vector. This allows later to find the closest vector of the opposite sentiment.

\begin{table}
\begin{center}
\begin{tabular}{l | c}
Model & Ratio changed \\
\hline
All phrases & 50.8 \\
Phrases longer than two words & 52.5 \\
Phrases longer than five words & 53.0
\end{tabular}
\end{center}
\caption{Success rate of the autoencoder in changing the sentiment of phrases}
\label{tab:autenc}
\end{table}

\begin{table*}[h]
\begin{center}
\scalebox{0.9}{
\begin{tabular}{l|l|l}
Original sequence & Generated sequence & Sentiment change\\
\hline \hline
no movement , no yuks , not much of anything & no movement , no yuks , not much of anything	 & no\\
this is one of polanski 's best films & this is one of polanski 's lowest films & yes\\
most new movies have a bright sheen & most new movies a unhappy moments & yes w/error \\
gollum 's ` performance ' is incredible & gollum 's ` performance ' not well received ! & yes\\
as a singular character study , it 's perfect & as a give study , it 's perfect & no w/error\\

\end{tabular}}
\caption{Examples of transformed sentences generated by the encoder-decoder model}
\label{tab:sents}
\end{center}
\end{table*}

\begin{table*}[h]
\begin{center}
\scalebox{0.9}{
\begin{tabular}{l|l|l}
Original sequence & Generated sequence & Sentiment change\\
\hline \hline
no movement , no yuks , not much of anything & obvious movement, obvious yuks, much of kind & no w/error\\
this is one of polanski 's best films & this is one of polanski 's worst films & yes\\
most new movies have a bright sheen & most new movies have a bleak ooze & yes w/error\\
gollum 's ` performance ' is incredible & gollum 's ` performance ' is unbelievable & undefined 
 \\
as a singular character study , it 's perfect & as a singular character examination it 's crisp & yes w/error \\
\end{tabular}}
\caption{Examples of transformed sentences (same as Table \ref{tab:sents}) using the word vectors approach}
\label{tab:sents2}
\end{center}
\end{table*}

Table \ref{tab:autenc} shows the success rate of the autoencoder in terms of changing the sentiment of the phrases extracted by the attention mechanism. After being decoded, 50.8 percent of the phrases are classified as a different sentiment from the one they originally were classified. The number reported is the ratio of sentences that got assigned a different sentiment by the sentiment classifier after the transformation. Furthermore, some of the phrases in the extracted set had a length of only one or two words for which it is hard to predict the sentiment. These short sequences were included because in the final model they would also be extracted, so they do have an impact on the performance. The model was also tested while leaving out the shorter phrases, both on phrases longer than two and longer than five words, which slightly increases the success rate. 

\subsection{Full model}
The full pipeline was tested in two ways. First, sentences were evaluated using a group of human evaluators to determine whether the sentences generated were grammatically and semantically correct on top of the change in sentiment. Next, the change in sentiment was tested using the sentiment classifier described by \cite{yang2016hierarchical}. 

To find how well the full pipeline performed in changing the sentiment of sequences, a basic human evaluation was performed (as a first experiment) on a subset of generated sequences based on sentences from the rotten tomatoes dataset \cite{pang2005seeing}. The reason for choosing this dataset is that the sentences were shorter, so the readability was better than using the imdb dataset. The setup was as follows: Reviewers were shown the original sentence and the two variants generated by two versions of the algorithm, the encoder-decoder model and the word vector model. Reviewers were then asked to rate the generated sentence on a scale from 1 to 5, both in terms of grammatical and semantic correctness and the extent to which the sentiment had changed. The rating of grammatical and semantic correctness was so that the reviewers could indicate whether a sentence was still intelligible after the change was performed. The rating of the sentiment change was an indication of how much the sentiment changed towards the opposite sentiment. In this case, a perfect change of sentiment from positive to negative (or vice-versa) would be rated as 5 and the sentiment remaining exactly the same would be rated as 1. Reviewers also had the option to mark that a sentence hadn't changed, as that would not change the sentiment but give a perfect score in correctness. After all reviewers had reviewed all sentences, the average score for both correctness and sentiment change was calculated for both approaches. The number of times a sentence hadn't changed was also reported. The two approaches were then compared to see which approach performed better.

Tables \ref{tab:sents} and \ref{tab:sents2} show how some sentences are transformed using the encoder-decoder and the word vectors approach respectively, along with information on whether the sentiment was changed (and if that happened with introducing some grammatical or semantic error). The word vectors approach seems to do a better job at replacing words correctly, however in both cases there are some errors which are being introduced.

Table \ref{tab:quest} shows the results obtained by the human evaluation. The numbers at grammatical correctness and sentiment change are the average ratings that sentences got by the evaluation panel. Last row shows the percentage of sentences that did not change at all. The test group indicated that the encoder approach changed the sentence in slightly more than 60\% the cases, while the word vectors approach did change the sentence in more than 90\% of the cases. This is possibly caused by the number of unknown tokens in the sentences, which caused problems for the encoder, but not for the word vector approach, as it would just ignore the unknown tokens and move on. Another explanation for this result is that the attention mechanism only highlights single words and  without the help of an emotion lexicon these single replacements often do not change the sentiment of the sentence, as can be seen in Table \ref{tab:sents}. 

Table \ref{tab:quest} also shows that the grammatical quality of the sentences and the sentiment change as performed by the word vectors approach was evaluated to be higher than the ones generated by the encoder approach. Observing the changes made to sentences shows that the replacements in the word vector approach were more sensible when it comes to word type and sentiment. The cause of this is that the word vector approach makes use of an emotion lexicon, which ensures that each word inserted is of the desired sentiment. The encoder approach makes use of the fixed-word vector and the sentiment as determined by the sentiment classifier of the whole encoded phrase, allowing for less control on the exact sentiment of the inserted phrase.

\begin{table}[h]
\begin{center}
\begin{tabular}{l | c | c}
Question & Encoder  & Word vectors\\
\hline \hline
Grammatical correctness & $2.7/5$ & $4.4/5$\\
Sentiment change & $3.5/5$ & $4.3/5$ \\
\hline
Unchanged & $36.67\%$ & $6.67\%$
\end{tabular}
\caption{Average score one a scale from 1 to 5 for correctness and sentiment change reviewers assigned to the sentences and ratio of sentences that remained unchanged.}
\label{tab:quest}
\end{center}
\end{table}

The second experiment conducted had the goal to test the ratio of sentences that changed sentiment compared to the original one. This model is also better able to give an objective measure on how well the model does what it is supposed to do, namely changing the sentiment.

\begin{table}[h]
\begin{center}
\begin{tabular}{l || c | c}
Model & Rotten Tomatoes & IMDB\\ 
\hline
Decoder & 53.6 & 53.7 \\
Word vectors & 49.1 & 53.3
\end{tabular}
\caption{Percentage of sentences that changed sentiment}
\label{tab:accuracy}
\end{center}
\end{table}

Table \ref{tab:accuracy} shows that the accuracy in changing the sentiment is by around 5\% higher on the rotten tomatoes coprus \cite{pang2005seeing} but similar for the imdb corpus \cite{maas-EtAl:2011:ACL-HLT2011}. It should be noted that the performance of the encoder-decoder is almost identical for both datasets.
 
\section{\uppercase{Discussion}}
The model proposed in this paper transforms the sentiment of a sentence by replacing short phrases that determine the sentiment. Extraction of these phrases is done using a sentiment classifier with an attention mechanism. These phrases are then encoded using an encoder-decoder network that is trained on these phrases. After the phrases are encoded, the closest phrase of the opposite sentiment is found and replaced into the original sentence. Alternatively, the extracted phrase is transformed by finding the closest word of the opposite sentiment using an emotion lexicon to assign sentiment to words.

The model was evaluated on both its individual parts and end-to-end. 
We used both automatic metrics and human evaluation. Testing the success rate (of changing the sentiment), best results were achieved with the encoder-decoder method, which score more than 50 \% on both datasets. Human evaluation on the model gave the best scores to the word vector based model, both in terms of the change of sentiment and in terms of the grammatical and semantic correctness.

Results raise the issue of language interpretability by humans and machines. Our method seems to create samples that are sufficiently changing the sentiment for the classifier (thus the goal of creating new data points is successful), however this is not confirmed by the human evaluators who judge the actual content of the sentence. However, it should be noted here that human evaluation experiments need to be extended once the approach is more robust to confirm the results.

As for future work, we plan to introduce a more carefully assembled dataset for the encoder-decoder approach, since that might improve the quality of the decoder output. The prominence of unknown tokens in the data suggests that experimenting with a character-level implementation might improve the results, as such algorithms can often infer the meaning of all words, regardless of how often they appear in the data. This could solve the problem of not all words being present in the vocabulary which results in many unknown tokens in the generated sentences. 

Finally, another way to improve the model is to have the encoder-decoder better caption the phrases in the latent space. We based our model on \cite{cho2014learning} but used less hidden units (due to hardware limitations) which may have caused learning a worse representation of the phrases in the latent space. Using more hidden units (or a different architecture for the encoder/decoder model) is a way to further explore how reuslts could be improved. 

\bibliographystyle{apalike}
{\small
\bibliography{icaart2018}
}\end{document}